\documentclass[10pt, a4paper]{scrartcl}

\usepackage{latexsym}
\usepackage{amssymb}
\usepackage[nonamelimits]{amsmath}
\usepackage{url}

\usepackage{color}
\definecolor{myred}{rgb}{0.5,0,0}
\definecolor{myblue}{rgb}{0,0,0.75}
\definecolor{mygreen}{rgb}{0,0.5,0}

\usepackage{ifpdf}

\ifpdf
   \usepackage[pdftex]{graphicx}
   \usepackage[pdftex,
              pdftitle={Minimising quantifier variance under prior probability shift},
               pdfsubject={},
               pdfkeywords={},
               pdfauthor={Dirk Tasche},
               pdfstartview=FitR,
               breaklinks=true,
               colorlinks=true,
               citecolor=myred,
               linkcolor=myblue,
               urlcolor=mygreen]{hyperref}
\else
   \usepackage[dvips]{graphicx}
   \usepackage{hyperref}
   \usepackage{rotating}
\fi

\newtheorem{theorem}{Theorem}[section]

\newtheorem{proposition}[theorem]{Proposition}

\numberwithin{equation}{section}

\title{Minimising quantifier variance under prior probability
shift}

\author{%
Dirk Tasche\thanks{Independent researcher, e-mail: dirk.tasche@gmx.net}}

\date{}

\begin{document}

\maketitle

\begin{abstract}
 For the binary prevalence quantification problem under prior probability
shift, we determine the asymptotic variance of the maximum likelihood 
estimator. We find that it is a function of the Brier score for
the regression of the class label on the features under the test data set
distribution. This observation suggests that optimising the accuracy of
a base classifier, as measured by the Brier score, on the training data set helps to reduce the variance
of the related quantifier on the test data set. Therefore, we also point out training criteria
for the base classifier that imply optimisation of both of the Brier scores on the
training and the test data sets.
\end{abstract}

\section{Introduction}

The survey paper \cite{Gonzalez:2017:RQL:3145473.3117807} described the problem to estimate prior class
probabilities (also called prevalences) on a test set with a different distribution than the training set 
(the \emph{quantification} problem)
as ``Given a labelled training set, 
induce a quantifier that takes an unlabelled test set as input and
returns its best estimate of the class distribution.'' As becomes clear from
\cite{Gonzalez:2017:RQL:3145473.3117807} and also more recent work on the problem,
it has been widely investigated in the past twenty years. 

A lot of different approaches to quantification of prior class probabilities has
been proposed and analysed (see, e.g.\ \cite{Gonzalez:2017:RQL:3145473.3117807, 
hassan2020accurately, moreo2021quapy}), but it appears that the following 
question has not yet received very much
attention:

\emph{Is it worth the effort to try to train a good (accurate) hard (or soft or probabilistic)
classifier as the `base classifier' for the task of quantification if
the class labels of individual instances are unimportant and only
the aggregate prior class probabilities are of interest?}

In principle, there is a clear answer to this question. The accuracy of the
classifier matters at least in the extreme cases: 
\begin{itemize}
\item If a classifier is least accurate because its predictions and the
true class labels are stochastically independent, then quantification is not feasible.
\item If a classifier is most accurate in the sense of making perfect predictions then
perfect quantification is easy by applying Classify \& Count \cite{forman2005counting}.
\end{itemize}
But if no perfect classifier is around, can we be happy to deploy a moderately accurate classifier
for quantification or should we rather strive to develop an optimal classifier, possibly
based on an comprehensive feature selection process?

Some researchers indeed suggest that the accuracy of the
base classifiers is less important for quantification than for classification.
 \cite{forman2008quantifying} made the following statements:
\begin{itemize}
\item From the abstract of \cite{forman2008quantifying}: ``These strengths can make quantification practical
for business use, even where classification accuracy is poor.''
\item P.~165 of \cite{forman2008quantifying}: ``The effort to develop special 
purpose features or classifiers could increase the cost
significantly, with no guarantee of an accurate classifier. Thus, an imperfect classifier
is often all that is available.''
\item P.~166 of \cite{forman2008quantifying}: ``It is sufficient but not 
necessary to have a perfect classifier in order to estimate the
class distribution well. If the number of false positives balances against false negatives,
then the overall count of predicted positives is nonetheless correct. Intuitively,
the estimation task is easier for not having to deliver accurate predictions on individual
cases.''
\end{itemize}
The point on the mutual cancellation of false positives and false negatives is mentioned also by
a number of other researchers like for instance \cite{Esuli2010}. On p.~74, \cite{Esuli2010} wrote:
``Equation~1 [with the definition of the $F$-measure] shows that $F_1$ deteriorates
with $(FP + FN)$ and not with $|FP - FN|$, as would instead be required of a function that truly optimizes
quantification.''

There are also researchers that hold the contrary position, at least as quantification
under an assumption of prior probability shift (see \eqref{eq:shift} below for the formal 
definition) is concerned: 
\begin{itemize}
\item \cite{vaz2017prior} noted for the class of `ratio estimators' they introducted that
it was both desirable and feasible to construct estimators with small asymptotic variances.
\item \cite{tasche2019confidence} demonstrated by a simulation study that estimating 
the prior class probabilities by means of a more accurate base classifier may entail
much shorter confidence intervals for the estimates.
\item \cite{alexandari2020maximum, Garg2020Unified} pointed out the efficiency of the 
maximum likelihood estimator (MLE)
for the quantification task and made cases for its application.
\end{itemize}
In the following, we revisit the question of the usefulness of accurate classifiers for
quantification:

After giving an overview of related research in Section~\ref{se:work} and specifying 
the setting and assumptions for the binary quantification problem 
in Section~\ref{se:setting}, we recall the technical details of
the definition of the MLE for the positive class prevalence in Section~\ref{se:ML}. In particular,
we show that the MLE is well-defined under the mild condition that the test sample consists of at least 
two different points, see \eqref{eq:c2} below.

In Section~\ref{se:Cramer}, we describe, based on its representation in terms of the Fisher information,  
the Cram\'er-Rao lower bound for the
variances of unbiased estimators of the positive class prevalence, see \eqref{eq:bound} below. 
This lower bound, at the same time, is the large sample variance of the MLE defined in Section~\ref{se:ML}.
Thus, the Cram\'er-Rao lower bound is achievable in theory by MLEs -- which might explain to
some extent the superiority of MLEs as observed in \cite{alexandari2020maximum, Garg2020Unified}.

In Section~\ref{se:Brier}, we show for the test distribution of the feature vector
that its associated Fisher information with respect to the positive
class prevalence is -- up to a factor depending only on the true prevalence -- just the variance
of the posterior positive class probability (Proposition~\ref{pr:Fisher} below). The variance
of the posterior probability is closely related to the Brier score for the regression of 
the class label on the feature vector. While the Brier score decreases when the information content
of the feature vector increases, the variance of the posterior probability decreases with shrinking 
information content of the features. 
In any case, these observations imply that the large sample variance of the MLE (or 
the Cram\'er-Rao lower bound for the variances of unbiased estimators) may be reduced when
a feature vector with larger information content is selected. Section~\ref{se:Brier} concludes
with two suggestions of how this can be achieved in practice (Brier curves, ROC analysis).

In Section~\ref{se:example}, we illustrate the observations of Sections~\ref{se:Cramer} and
\ref{se:Brier} with a numerical example. Table~\ref{tab:1} below demonstrates variance reduction
through more powerful features both for the ML quantifier and a non-ML quantifier.
Figure~\ref{fig:1} below suggests that differences in efficiency between the ML quantifier and
non-ML quantifiers may depend both on the information content (power) of the feature vector and
on the true value of the positive class prevalence.

\section{Related work}
\label{se:work}

Prior probability shift is a special type of data set shift, see \cite{MorenoTorres2012521}
for background information and a taxonomy of data set shift.
In the literature, also other terms are used for prior probability shift, 
for instance `global drift' \cite{hofer2013drift} or `label shift' \cite{pmlr-v80-lipton18a}.

The problem of estimating the test set prior class probabilities can
also be interpreted as a problem to estimate the parameters of a `mixture model' 
\cite{fruhwirth2006finite} where the component distributions are learnt on 
a training set. See \cite{peters1976numerical} for an early work on 
the properties of the maximum likelihood (ML) estimator in this case.
\cite{saerens2002adjusting} revived the interest in the ML estimator for
the unknown prior class probabilities in the test set by specifying the
associated `expectation maximisation' (EM) algorithm. 

\cite{tasche2017fisher} proposed to take recourse to the notion
of Fisher consistency as a criterion to identify completely unsuitable approaches to the
quantification problem that do not have this property. 
\cite{tasche2017fisher} then proved Fisher consistency of the ML estimator
under prior probability shift.

The ML approach has been criticised for its sometimes moderate performance
and the effort and amount of training data needed to implement it. However,
recently some researchers \cite{alexandari2020maximum, Garg2020Unified} began to vindicate the
ML approach. They focussed on the need to properly calibrate the posterior class
probability estimate in order to improve the efficiency of the MLE for the positive 
class prevalence on the test set. Complementing the work of \cite{alexandari2020maximum, Garg2020Unified},
in this paper we study the role that constructing more powerful (or accurate) classifiers 
by selection of more appropriate features may play to reduce the variances of the
related quantifiers.

In the following, we revisit the well-known asymptotic efficiency property of
ML estimators in the special case of the MLE for prior class probabilities
in the binary setting and investigate how it is impacted by the power (accuracy) of 
the classifier on the MLE is based.

\section{Setting}
\label{se:setting}

We consider the binary prevalence quantification problem in the following setting:
\begin{itemize}
\item There is a training (or source) data set $(x_1, y_1), \ldots, (x_m, y_m) \in 
\mathfrak{X}\times \{0,1\}$. It is assumed to be an i.i.d.\ sample of a random vector $(X, Y)$ with values in 
$\mathfrak{X}\times \{0,1\}$. The vector $(X,Y)$ is defined on a probability space $(\Omega, P)$,
the training (or source) domain.
The elements $\omega$ of $\Omega$ are the instances (or objects). Each instance $\omega$ belongs to one
of the classes $0$ and $1$, and its class label is $Y(\omega) \in \{0,1\}$. In addition, each
instance $\omega$ has features $X(\omega) \in \mathfrak{X}$. Often, $\mathfrak{X}$ is 
the $d$-dimensional Euclidian space such that accordingly $X$ is a real-valued random vector.
See Appendix~B.1 of \cite{zhao2013beyond} for more detailed comments of how this
setting avoids the logical problems that arise when feature vectors and instances are
considered to be the same thing.
\item Under the training distribution $P$, both the features $X(\omega)$ and the class labels $Y(\omega)$
of the instances are observed
in a series of $m$ independent experiments resulting in the sample $(x_1, y_1), \ldots, (x_m, y_m)$.
The sample can be used to infer the joint distribution of $X$ and $Y$ under $P$, and hence, 
in particular, also the distribution of $Y$ (the class distribution) under $P$.
\item There is a test (or target) data set $z_1, \ldots, z_n \in \mathfrak{X}$. 
It is assumed to be an i.i.d. sample of the random vector $X$ with values in $\mathfrak{X}$,
under a probability measure $Q$ on $\Omega$ that may be different to the training distribution $P$.
\item Under the test distribution $Q$, only the features $X(\omega)$ of the instances are observed 
in a series of $n$ independent experiments resulting in the sample $z_1, \ldots, z_n$. The sample
can be used to infer the distribution of $X$ under $Q$.
\item The goal of quantification is to infer the distribution of $Y$ under $Q$, based 
on the sample of features $z_1, \ldots, z_n$ generated under $Q$ and on the joint sample of features
and class labels $(x_1, y_1), \ldots, (x_m, y_m)$ generated under $P$. It is not possible 
to design a method for this inference without any assumption on the relation of $P$ and $Q$.
\item In this paper, we assume that $P$ and $Q$ are related by \emph{prior probability shift}, in
the sense that the class-conditional feature distributions are the same under $P$ and $Q$, i.e.\ it
holds that
\begin{equation}\label{eq:shift}
P[X\in M\,|\,Y=i]\ = \ Q[X\in M\,|\,Y=i]
\end{equation}
for $i \in \{0,1\}$ and all measurable subsets $M$ of $\mathfrak{X}$.
\end{itemize}

Denoting $P[Y=1] = p$ and $Q[Y=1] =q$, \eqref{eq:shift} implies that the distribution 
of the features $X$ under $P$ and $Q$ respectively can be represented as
\begin{subequations}
\begin{align}
P[X \in M] & =
p\,P[X\in M\,|\,Y=1] + (1-p)\,P[X \in M\,|\,Y=0], \\
Q[X \in M] & = \label{eq:uncond} 
q\,P[X\in M\,|\,Y=1] + (1-q)\,P[X \in M\,|\,Y=0],
\end{align}
\end{subequations}
for $M \subset \mathfrak{X}$. In the following, we assume that the components
$p$, $P[X\in M\,|\,Y=1]$ and $P[X\in M\,|\,Y=0]$ can be perfectly estimated
from the training sample $(x_1, y_1), \ldots, (x_m, y_m)$. 

Basically, this means letting $m= \infty$ which obviously is infeasible. The assumption helps,
however, to shed light on the importance of both maximum likelihood estimation and
accurate classifiers for the efficient estimation of the unknown positive class prevalence
$q$ in the test data set.

\section{The ML estimator for the positive class prevalence}
\label{se:ML}

Assume that the conditional distributions in \eqref{eq:shift}
have positive densities $f_i$, $i= 0, 1$. Then the unconditional density of the features
vector $X$ under $Q$ is
\begin{equation}\label{eq:uncond.dens}
f^{(q)}(x) \ = \ (1-q)\,f_0(x) + q\,f_1(x), \quad x \in \mathfrak{X}.
\end{equation}
Hence the likelihood function 
\begin{equation*}
L_n(q) = L_n(q; z_1, \ldots, z_n) 
\end{equation*}
for the sample $z_1, \ldots, z_n$ is given by 
\begin{equation}\label{eq:Ln}
L_n(q) \ = \ \prod_{i=1}^n \bigl(q\,(f_1(z_i)-f_0(z_i)) + f_0(z_i)\bigr).
\end{equation}
This implies for the first two derivatives of the log-likelihood with respect to $q$:
\begin{subequations}
\begin{align}
\frac{\partial \log L_n}{\partial\, q}(q) & = 
    \sum_{i=1}^n \frac{f_1(z_i)-f_0(z_i)}{q\,(f_1(z_i)-f_0(z_i)) + f_0(z_i)},\\
\frac{\partial^2 \log L_n}{\partial\, q^2}(q) & = 
    - \sum_{i=1}^n \left(\frac{f_1(z_i)-f_0(z_i)}{q\,(f_1(z_i)-f_0(z_i)) + f_0(z_i)}\right)^2 
    \le 0.
\end{align}
\end{subequations}
We assume that  there is at least one $j \in \{1, \ldots, n\}$ such that
\begin{equation}\label{eq:c2}
f_1(z_j)\ \neq\ f_0(z_j).
\end{equation}
Under \eqref{eq:c2}, $q \mapsto \log L_n(q)$ is strictly concave in $[0,1]$.
Hence (see Example~4.3.1 of \cite{titterington1985statistical}) the equation
\begin{equation}\label{eq:first}
\frac{\partial \log L_n}{\partial\, q}(q) \ = \ 0
\end{equation}
has a solution $0 < q < 1$ if and only if
\begin{subequations}
\begin{equation}\label{eq:c1}
\begin{split}
\frac{1}{n} \sum_{i=1}^n \frac{f_1(z_i)}{f_0(z_i)} > 1 \quad\text{and}\quad
\frac{1}{n} \sum_{i=1}^n \frac{f_0(z_i)}{f_1(z_i)} > 1.
\end{split}
\end{equation}
This solution is then the unique point in $[0,1]$ where $L_n(q)$ takes its absolute maximum value.
By strict concavity of $\log L_n$, if \eqref{eq:c1} is not true then
either 
\begin{equation}\label{eq:c3}
\begin{split}
\frac{1}{n} \sum_{i=1}^n \frac{f_1(z_i)}{f_0(z_i)} \le 1 \quad\text{and}\quad
\frac{1}{n} \sum_{i=1}^n \frac{f_0(z_i)}{f_1(z_i)} > 1,
\end{split}
\end{equation}
applies or
\begin{equation}\label{eq:c4}
\begin{split}
\frac{1}{n} \sum_{i=1}^n \frac{f_1(z_i)}{f_0(z_i)} > 1 \quad\text{and}\quad
\frac{1}{n} \sum_{i=1}^n \frac{f_0(z_i)}{f_1(z_i)} \le 1,
\end{split}
\end{equation}
holds. Under \eqref{eq:c3}, the unique maximum of $\log L_n$ in $[0,1]$ lies at $q=0$
while under \eqref{eq:c4}, the unique maximum of $\log L_n$ in $[0,1]$ is taken at $q=1$.
\end{subequations}

In summary, under the natural assumption \eqref{eq:c2}, the likelihood function $L_n$ 
of \eqref{eq:Ln} has an absolute maximum in $[0,1]$ at a unique point $q^\ast \in[0,1]$.
As a consequence,
the maximum likelihood (ML) estimate $\hat{q}_n$ of the test set positive class prevalence $q$
is well-defined by setting $\hat{q}_n = q^\ast$.

\section{The Cram\'er-Rao bound for unbiased estimators}
\label{se:Cramer}

In the setting of Section~\ref{se:setting}, let $\widetilde{q}_n$ be any unbiased 
estimator of the positive class prevalence $q$ under
the test distribution $Q$, i.e.\ $\widetilde{q}_n = W_n(X_1, \ldots, X_n)$ for some function
$W_n: \mathfrak{X}^n \to \mathbb{R}$ such that 
\begin{equation*}
E_Q[\widetilde{q}_n] = Q[Y=1] = q,
\end{equation*}
where $E_Q$ denotes the expected value under the mixture probability measure
$Q$ of \eqref{eq:uncond}. 
Assume additionally that there are positive densities $f_0$ and $f_1$ of the
class-conditional distributions of the feature vector $X$, such that the density $f^{(q)}$ of
$X$ under $Q$ is given by \eqref{eq:uncond.dens}.

If $(X_1, \ldots, X_n)$ is an  i.i.d.\ sample from the distribution of the feature vector under the
test distribution $Q$, then  the variance of $\widetilde{q}_n$ is bounded from below by the inverse of
the product of the Fisher information of the test distribution with respect to $q$ 
and the size of the sample (Cram\'er-Rao bound, see, e.g., 
Corollary~7.3.10 of \cite{Casella&Berger}):
\begin{align}
\mathrm{var}[\widetilde{q}_n] & \ \ge \ \frac{1}{n\,E_Q\left[
    \left(\frac{\partial \log f^{(q)}}{\partial\, q}(X)\right)^2\right]}\notag\\
    & \ =\ 
    \frac{1}{n\,E_Q\left[\left(
    \frac{f_1(X) - f_0(X)}{f^{(q)}(X)}\right)^2 \right]}.\label{eq:bound}
\end{align}

Actually, since $\hat{q}_n$ from Section~\ref{se:ML} is the ML estimator of $q$,
\eqref{eq:uncond} presents not only a lower bound for the variances of unbiased estimators of $q$ but
also the large sample variance of $\hat{q}_n$ in the sense that $\sqrt{n}\,(\hat{q}_n - q)$ converges 
in distribution toward $\mathcal{N}\left(0, E_Q\left[\Big(
    \frac{f_1(X) - f_0(X)}{f^{(q)}(X)}\Big)^2 \right]^{-1}\right)$, the normal
    distribution with mean $0$ and variance $E_Q\left[\Big(
    \frac{f_1(X) - f_0(X)}{f^{(q)}(X)}\Big)^2 \right]^{-1}$, the 
    \emph{asymptotic variance} of $\hat{q}_n$ (see, e.g., Theorem~10.1.12 of
 \cite{Casella&Berger}). Note that the lower bound of \eqref{eq:bound} may not hold for 
 $\hat{q}_n$ because it need not be unbiased.
 
In summary, if the densities of the class-conditional feature distributions are known, the ML estimator
of $q$ has asymptotically the smallest variance of all unbiased estimators of $q$ on i.i.d.\ samples 
from the test distribution of the features. This is demonstrated in the two upper panels 
of Table~4 of \cite{tasche2019confidence} which shows on simulated data 
that confidence intervals for $q$ -- which are
primarily driven by the standard deviations of the estimators -- based
on the ML estimator are the shortest if the training sample is infinite and the
test sample is large.

\section{The asymptotic variance of the ML estimator}
\label{se:Brier}
\label{se:asym}

Denote by $\eta_Q(x)$ the posterior positive class probability given $X=x$ under $Q$. 
Assume that the feature vector $X$ under $Q$ has a density that is given by \eqref{eq:uncond.dens}.
Then it holds that
\begin{equation}\label{eq:eta}
\eta_Q(x) \ =\ \frac{q\,f_1(x)}{f^{(q)}(x)}.
\end{equation}
This representation immediately implies the following result on the representation of the Fisher information
mentioned above in the context of the Cram\'er-Rao bound in terms of the variance of $\eta_Q$.
\begin{proposition}\label{pr:Fisher}
If the feature vector $X$ under $Q$ has a density as specified by \eqref{eq:uncond.dens}
then the Fisher information $E_Q\left[ \left(\frac{\partial \log f^{(q)}}{\partial\, q}(X)\right)^2\right]$ 
of the distribution of $X$ under $Q$ with respect to $q$ can be represented as follows:
$$E_Q\left[
    \left(\frac{\partial \log f^{(q)}}{\partial\, q}(X)\right)^2\right] =
    \frac{\mathrm{var}_Q[\eta_Q(X)]} {q^2\,(1-q)^2}.$$   
\end{proposition}
From Proposition~\ref{pr:Fisher} we obtain the following representation of the 
asymptotic variance of the ML estimator $\hat{q}_n$:
\begin{equation}
E_Q\left[\Big(\frac{f_1(X) - f_0(X)}{f^{(q)}(X)}\Big)^2 \right]^{-1} 
=
\frac{q^2\,(1-q)^2}{\mathrm{var}_Q[\eta_Q(X)]}.\label{eq:asym}
\end{equation}
Recall the following decomposition of the optimal Brier Score $BS_Q(X)$ for the problem
to predict the class variable $Y$ from the features $X$ (under the test distribution
$Q$):
\begin{align}
BS_Q(X) &\ = \ E_Q\left[(Y-\eta_Q(x))^2\right] \notag\\
& \ = \ \mathrm{var}_Q[Y] - \mathrm{var}_Q[\eta_Q(X)]\notag\\
& \ = \ q\,(1-q) - \mathrm{var}_Q[\eta_Q(X)].\label{eq:decomp}
\end{align}
In \eqref{eq:decomp}, the optimal Brier Score $BS_Q(X)$ is also called \emph{refinement loss},
while $\mathrm{var}_Q[Y]$ and $\mathrm{var}_Q[\eta_Q(X)]$ are known as \emph{uncertainty}
and \emph{resolution} respectively \cite{Hand97}.

By \eqref{eq:asym}, the asymptotic variance of the ML estimator $\hat{q}_n$ is
reduced if a feature vector $X$ with greater variance of $\eta_Q(X)$ is found (or if the Brier Score
with respect to $X$ decreases). Consider, for instance, a feature vector $X'$ that is a function 
of the feature vector $X$, i.e.\ it holds that $X' = F(X)$ for some function 
$F: \mathfrak{X}\to\mathfrak{X}'$. Since $F$ potentially maps different values $x \in \mathfrak{X}$
onto the same value $x' \in \mathfrak{X}'$ the amount of information carried by $X'$ is reduced
compared to the amount of information carried by $X$. Therefore, the approximation of the class label $Y$
by regression on $X'$ is less close than the approximation of $Y$ by regression on $X$.

From this observation, it follows that $BS_Q(X') \ge 
BS_Q(X)$. This in turn implies by \eqref{eq:decomp} and \eqref{eq:asym} for the 
asymptotic variances of the ML estimators $\hat{q}_n(X')$  and $\hat{q}_n(X)$ that
\begin{equation}\label{eq:varQ}
\mathrm{var}_Q[\hat{q}_n(X')] \ \le \  \mathrm{var}_Q[\hat{q}_n(X)].
\end{equation}
Observe that $X' = F(X)$ also implies $BS_P(X') \ge 
BS_P(X)$, i.e.\ also under the training distribution $P$, the posterior positive class
probability $\eta_P(X)$ based on $X$ is a better predictor
of $Y$ than $\eta_P(X')$ which is based on $X'$. By the assumption underlying this paper, $BS_P(X')$ and 
$BS_P(X)$ are observable while $BS_Q(X')$ and 
$BS_Q(X)$ are not, because the class label $Y$ is not observed in the test data set.

But does $BS_P(X') \ge BS_P(X)$ always imply $BS_Q(X') \ge BS_Q(X)$ and therefore also \eqref{eq:varQ},
thus generalising the implication ``$X' = F(X)$ $\Rightarrow$ \eqref{eq:varQ}''?

This paper has no fully general answer to this question. Instead we can only point to alternative
conditions on $X'$ and $X$ that imply both $BS_P(X') \ge BS_P(X)$ and $BS_Q(X') \ge BS_Q(X)$,
but are weaker than $X' = F(X)$.
\begin{description}
\item[Brier curves:] Recall the notion of \emph{Brier curve} from \cite{hernandez2011brier} (with the slightly
modified definition of \cite{tasche2021calibrating}). If the Brier curve for $\eta_P(X')$ dominates
the Brier curve for $\eta_P(X)$, then by item 6) of Proposition~5.2 and 
Proposition~4.1 of \cite{tasche2021calibrating},
it follows that $BS_P(X') \ge BS_P(X)$ and $BS_Q(X') \ge BS_Q(X)$ hold.
\item[ROC analysis:] Recall the notion of \emph{Receiver Operating Characteristic (ROC)} as defined, 
for instance, in \cite{reid2011information}. If the ROC for the density ratio associated with
$X$ dominates the ROC for the density ratio associated with $X'$, then by Remark~5.4 and 
Proposition~4.1 of \cite{tasche2021calibrating}, it follows that 
$BS_P(X') \ge BS_P(X)$ and $BS_Q(X') \ge BS_Q(X)$ hold.
\end{description}

\section{Example: Binormal model}
\label{se:example}

In this section, we numerically compare the variance of the 
Sample Mean Matching (SMM) estimator $\hat{q}_{n, SMM}$
of class prevalences \cite{hassan2020accurately} and 
the Cram\'er-Rao bound of \eqref{eq:bound} (which is also
the large sample variance of the ML estimator $\hat{q}_{n, ML}$ as specified
in Section~\ref{se:ML} above). In order to be able to do this,
we take recourse to the univariate binormal model with equal variances of the class-conditional
distributions: The two normal class-conditional distributions of the feature variable $X$ are given
by
\begin{subequations}
\begin{equation}\label{eq:binorm}
X\,|\,Y=i \ \sim \ \mathcal{N}(\mu_i, \sigma^2), \qquad i=0,1,
\end{equation}
for conditional means $\mu_0 < \mu_1$ and some $\sigma > 0$. For the sake of simplicity,
we choose 
\begin{equation}\label{eq:spec}
\mu_0 = 0, \quad \mu_1 > 0, \quad \sigma = 1.
\end{equation}
\end{subequations}
Greater values of $\mu_1$ imply less overlap of the class-conditional feature distributions, corresponding
to more powerful (or accurate) models. Or in other words, for greater values of $\mu_1$, 
the feature variable $X$ carries more information on the class label $Y$.

As stated in Section~\ref{se:setting}, we assume we are dealing with 
an infinitely large training sample and a test sample of size $n$.
By Section~\ref{se:asym}, then for large $n$ the variance of $\hat{q}_{n, ML}$
is approximately 
\begin{equation}\label{eq:varML}
\frac{1}{n}\,E_Q\left[\left(\frac{f_1(X) - f_0(X)}{f^{(q)}(X)}\right)^2 \right]^{-1},
\end{equation}
where $Q$ denotes the distribution underlying the test sample and $f_0$ and $f_1$
are the class-conditional feature densities -- which are given for the purpose of
this section by \eqref{eq:binorm}. We evaluate the term given by \eqref{eq:varML} by 
means of one-dimensional numerical integration, making use of the R-function `integrate'
\cite{RSoftware2019}.

By Eq.~(2) of \cite{hassan2020accurately}, in the setting of
this paper as specified in Section~\ref{se:setting} above, the estimator $\hat{q}_{n, SMM}$
is given by the following explicit formula:
\begin{equation}\label{eq:SMMmean}
\begin{split}
\hat{q}_{n, SMM} &\ = \ \frac{\frac{1}{n}\sum_{i=1}^n X_i - \mu_0}{\mu_1-\mu_0}\\
	& \ = \ \frac{1}{n\,\mu_1}\sum_{i=1}^n X_i.
\end{split}
\end{equation}
By \eqref{eq:SMMmean}, $\hat{q}_{n, SMM}$ is an unbiased estimator of the positive class prevalence $q$.
For the variance of $\hat{q}_{n, SMM}$, we can refer to Theorem~3 of \cite{vaz2017prior}, 
case $n_{tr} = \infty$ in the notation of \cite{vaz2017prior}.
Observe that in the case of SMM and $n_{tr} = \infty$, it holds that the representation 
of the variance is exact, not only approximate. Hence we obtain
\begin{equation}\label{eq:varSMM}
\begin{split}
\mathrm{var}_Q[\hat{q}_{n, SMM}] & \ = \ \frac{\sigma^2 + 
	q\,(1-q)\,(\mu_0^2 + \mu_1^2)}{n\,(\mu_1-\mu_0)^2}\\
	& \ = \ \frac{1}{n} \left(\frac{1}{\mu_1^2} + q\,(1-q)\right).
\end{split}
\end{equation}
In the model specified by \eqref{eq:binorm} and \eqref{eq:spec}, the classification power (or accuracy)
is driven by the difference of the conditional means, i.e.\ by the mean conditional on the
positive class $\mu_1$. If
we measure the power by the Area under the Curve (AUC, see for instance
Section~6.1 of \cite{reid2011information}) in order to obtain a measure which
is independent of the class prevalences, AUC is a simple function of $\mu_1 = \mu_1 - \mu_0$:
\begin{equation}\label{eq:AUC}
AUC(\mu_1) \ = \ \Phi\left(\frac{\mu_1}{\sqrt{2}}\right),
\end{equation}
with $\Phi$ denoting the standard normal distribution function.
\begin{table}
\begin{center}
\caption{Illustration of the relations of
model power, standard deviation of the SMM quantifier and large
sample standard deviation of the ML quantifier. Sample size is 100, 
positive class prevalence in test set is 0.2. The parameter $\mu_1$ is the
expected sample mean conditional on the positive class, see \eqref{eq:binorm} and
\eqref{eq:spec}.}\label{tab:1}
\begin{tabular}{|r|r|r|r|}\hline
\multicolumn{1}{|c|}{$\mu_1$} & \multicolumn{1}{|c|}{$AUC(\mu_1)$} & 
\multicolumn{1}{|c|}{$\sigma_{SMM}$}  &
\multicolumn{1}{|c|}{$\sigma_{ML}$} \\ \hline \hline
 0.01 & 0.5028 & 10.0001 & 10.0000 \\ \hline
 0.05 & 0.5141 &  2.0004 & 2.0000 \\ \hline
  0.10 &  0.5282 &  1.0008 &  0.9999 \\ \hline
 0.25 & 0.5702 &  0.4020 & 0.3998 \\ \hline
 0.50 & 0.6382 &  0.2040 & 0.2000\\ \hline
  1.00 & 0.7602 &  0.1077 & 0.1017 \\ \hline
  1.50 & 0.8556 &  0.0777 & 0.0710 \\ \hline
 2.00 & 0.9214 &  0.0640 & 0.0571\\ \hline
 2.50 & 0.9615 &  0.0566 & 0.0498\\ \hline
3.00 &  0.9831 &  0.0521 & 0.0456\\ \hline
3.50 & 0.9933 &  0.0492 & 0.0432\\ \hline
4.00 & 0.9977 &  0.0472 & 0.0418\\ \hline
5.00 & 0.9998 &  0.0447 & 0.0405 \\ \hline
\end{tabular}
\end{center} 
\end{table}

From \eqref{eq:AUC} und \eqref{eq:varSMM}, it is clear that for fixed 
test sample size $n$ the variance of $\hat{q}_{n, SMM}$ decreases when the power
of the model increases. This is less obvious from \eqref{eq:varML} for the asymptotic
variance of $\hat{q}_{n, ML}$ but if follows from \eqref{eq:asym} in that case. 

Table~\ref{tab:1} above illustrates these observations. In the table, we use the 
notation
$$\sigma_{SMM} = \sqrt{\mathrm{var}_Q[\hat{q}_{n, SMM}]}$$ 
and
$$\sigma_{ML}  =  \sqrt{\mathrm{var}_Q[\hat{q}_{n, ML}]}.$$
Table~\ref{tab:1} suggests the following observations:
\begin{itemize}
\item At low levels of model power, small increases of power entail huge reductions of both the
SMM variance and the ML large sample variance.
\item The variance reductions become moderate or even low for moderate and high levels of 
model power (higher than 75\% AUC).
\item The efficiency deficiency of the SMM estimator (as expressed by its variance) 
compared to the large sample ML variance varies and is much larger for higher
levels of model power.
\end{itemize}
Figure~\ref{fig:1} below shows that the efficiency gain of the ML estimator compared to the
SMM estimator -- assuming an infinitely large training sample -- does not only
depend on the model power but also on the positive class prevalence.

Indeed, Figure~\ref{fig:1} suggests that a large efficiency gain is possible in the 
presence of a large difference in the prevalences of the two classes while the gain
is rather moderate in the case of almost equal class prevalences.
\begin{figure}
  \centering
  \includegraphics[width=\linewidth]{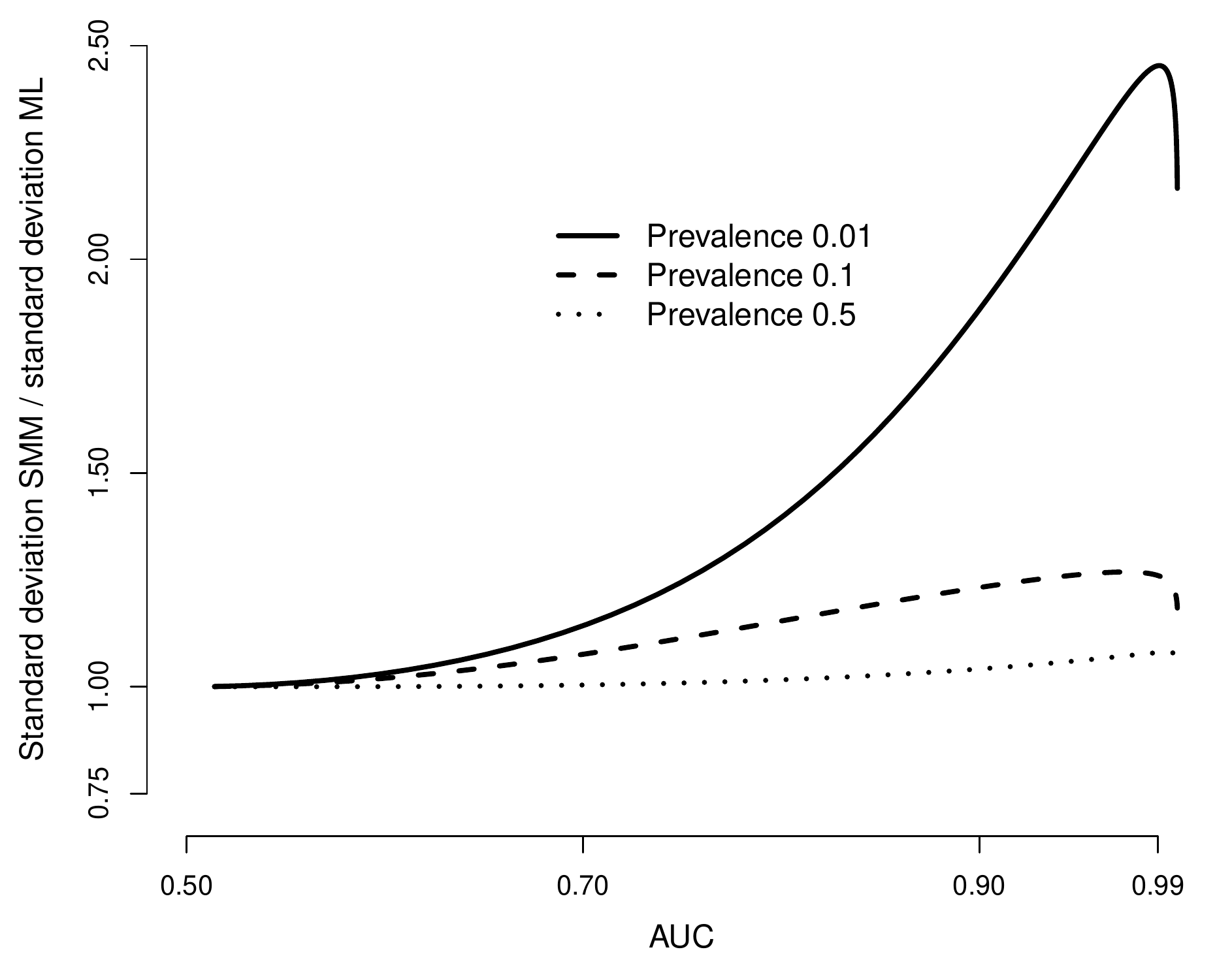}
  \caption{Ratios of standard deviations of SMM and ML estimates 
  as function of AUC and positive class prevalence.}\label{fig:1}
\end{figure}

\section{Conclusions}

In this paper, we have revisited the binary quantification problem, i.e.\ the problem of estimating
a binary prior class distribution on the test data set when training and test distributions are different. 
\begin{itemize}
\item Specifically, under the assumption of prior probability shift we have looked at the
asymptotic variance of the maximum likelihood estimator (MLE). 
\item We have found that this
asymptotic variance is closely related to the Brier score for the regression of the
class label variable $Y$ against the features vector $X$ under the test set distribution.
In particular, the asymptotic variance can be reduced by selection of a more powerful feature vector.
\item At the end of Section~\ref{se:Brier}, we have pointed out sufficient conditions and associated 
training criteria (Brier curves and ROC analysis) for
minimising both the Brier score on the training data set and the Brier score on the test data set.
\item These findings suggest methods to reduce the variance of the ML estimator of the prior class 
probabilities (or prevalences)
on the test data set. Due to the statistical consistency of ML estimators, by reducing the
variance of the estimator also its mean squared error is minimised.
\item The large sample variance of the MLE associated with its asymptotic variance is
identical to the Cram\'er-Rao lower bound for the variances of unbiased estimators of the
prior positive class probability. Therefore, it seems likely that also other estimators benefit
from improved performance of the underlying classifiers or feature vectors.
\end{itemize}
Indeed, results of a simulation study 
in \cite{tasche2019confidence} and theoretical findings in 
\cite{vaz2017prior, Vaz&Izbicki&Stern2019} suggest that improving the accuracy of the base
classifiers used for quantification helps to reduce not only the variances of ML estimators but
also of other estimators. The example of the Sample Mean Match (SMM) estimator we have discussed in 
Section~\ref{se:example} supports this conclusion.

However, these findings must be qualified in so far as the observations made in this paper
apply only to the case where both training data set and test data set are large. This is a 
severe restriction indeed as \cite{forman2008quantifying} pointed out the importance of
quantification methods specifically in the case of small training data sets, for cost efficiency
reasons.

Further research on developing efficient quantifiers for small or moderate training and test data set sizes
therefore is highly desirable. A promising step in this direction has already been done by
 \cite{maletzke2020importance}, with a proposal for the selection of the most suitable quantifiers 
for problems on data sets with widely varying sizes.


\end{document}